\documentclass[runningheads]{llncs}

\pdfoutput=1
\usepackage[T1]{fontenc}
\usepackage{graphicx}
\usepackage{hyperref}

\usepackage{algorithm}
\usepackage{algpseudocode}
\usepackage{amsmath}
\usepackage{amssymb}

\usepackage{color}

\begin{document}
\title{SimuDICE: Offline Policy Optimization Through World Model Updates and DICE Estimation}
\titlerunning{SimuDICE}
\author{Catalin E. Brita\inst{1} \and
Stephan Bongers\inst{1} \and
Frans A. Oliehoek\inst{1}}
\authorrunning{C.E. Brita et al.}
\institute{Delft University of Technology\\
\email{\{c.e.brita,s.r.bongers,f.a.oliehoek\}@tudelft.nl}}
\maketitle           
\begin{abstract}
In offline reinforcement learning, deriving an effective policy from a pre-collected set of experiences is challenging due to the distribution mismatch between the \textit{target policy} and the \textit{behavioral policy} used to collect the data, as well as the limited sample size. Model-based \mbox{reinforcement} learning improves sample efficiency by generating simulated experiences using a learned dynamic model of the environment. However, these synthetic experiences often suffer from the same distribution mismatch. To address these challenges, we introduce \mbox{SimuDICE}, a framework that iteratively refines the initial policy derived from offline data using synthetically generated experiences from the world model. \mbox{SimuDICE} enhances the quality of these simulated experiences by adjusting the sampling probabilities of state-action pairs based on \mbox{stationary} DIstribution Correction Estimation (DICE) and the estimated confidence in the model's predictions. This approach guides policy improvement by balancing experiences similar to those frequently encountered with ones that have a distribution mismatch. Our experiments show that SimuDICE achieves performance comparable to existing algorithms while requiring fewer pre-collected experiences and planning steps, and it remains robust across varying data collection policies.

\end{abstract}

\section{Introduction}

Reinforcement Learning (RL)~\cite{sutton2018reinforcement} has demonstrated numerous successes in domains such as games~\cite{hafner2023mastering} and robotics~\cite{tobin2017domain}, largely due to simulation-based trial and error~\cite{mnih2015humanlevel,silver2016mastering}. While feasible in game environments (where the game can be considered a simulator) or in simple real-world scenarios that can be accurately simulated, such direct or easy access to the environment is often not possible. Furthermore, in some areas such as medicine~\cite{murphy2001marginal} the deployment of a new policy, even just for the sake of performance evaluation, may be risky or costly.

Offline RL~\cite{levine2020OfflineRL}, also known as batch RL~\cite{lange2012}, addresses the challenge of training agents from static, pre-collected datasets, which are typically composed of off-policy data~\cite{sutton2018reinforcement}. A key issue in this context is the mismatch between the state visitation distribution of the \textit{target (candidate) policy} and the \textit{behavioral (logging) policy}. The problem is exacerbated throughout the learning process as this mismatch grows, potentially leading to issues such as divergence in off-policy learning~\cite{baird1995residual,tsitsiklis1996analysis}, making direct online-to-offline transitions challenging~\cite{fujimoto19a,kumar2019stabilizing}.

The agent has to perform \textit{offline policy evaluation} during the learning process, a task that is particularly challenging due to the policy-induced state-action distribution mismatch. Early work on this problem tackled the challenge using products of importance sampling~\cite{precup2001off}. Subsequent approaches have focused on improving the variance by directly learning density ratios~\cite{hallak2017consistent,liu2018breaking}. The DIstribution Correction Estimation (DICE) family of algorithms~\cite{nachum2019dualdice,yang2020offpolicy,zhang2020gendice,zhang2020gradientdice} achieves impressive results by leveraging optimization techniques to directly estimate stationary distribution corrections, significantly reducing variance.

Most prior work in offline RL consists of model-free methods. These studies show that directly using off-policy RL algorithms yields poor results due to distribution mismatch and function approximation errors. To address this, various modifications were proposed, such as Q-network ensembles~\cite{fujimoto19a,wu2019behavior}, regularization towards the behavioral policy~\cite{jaques2019way,kumar_aviral2019,wu2019behavior}, and implicit Q-learning~\cite{kostrikov2021implicit}.

Model-based RL (MBRL) involves learning a dynamics model of the environment that can be used for policy improvement. MBRL has shown significant success in online learning, demonstrating excellent sample efficiency~\cite{hafner2019dreamer,hafner2023mastering,hafner2020mastering}. However, applying MBRL algorithms directly to offline datasets presents challenges due to the same distribution mismatch issue. Specifically, it is difficult to obtain a globally accurate model because the dataset may not cover the entire state-action space. Consequently, planning with a learned model without safeguards against model inaccuracy can lead to `hallucinated` states~\cite{jafferjee2020hallucinating} and `model exploitation`~\cite{clavera2018model,janner2019trust,kurutach2018model}, leading to the possibility of poor policy performance.

Most of the prior work in offline MBRL \cite{Cang2021,Swazinna2021} pre-train a one-step forward model via maximum likelihood estimation to be a simple mimic of the world and then uses it to improve the policy, without any change to the model. This results in an \textit{objective mismatch}, namely the objective function used for model training (accurately predicting the environment) is unrelated to its utilization (policy optimization). Recent works have identified \textit{objective mismatch} in the model training and utilization as problematic \cite{eysenbach2021mismatched,lambert2020objective}.

We introduce \textit{SimuDICE}\footnote{\url{https://github.com/Catalin-2002/SimuDICE}}, an algorithm that iteratively improves the target policy by adjusting the sampling probabilities within a world model. Unlike prior methods that focus solely on generating safe experiences, we extend that approach by integrating DICE estimations, usually used for offline policy evaluation. These estimations reveal how the reward distributions shift from the behavioral dataset to the target policy, which leads us to choose to explore those transitions further. Our experiments show that incorporating the DICE estimations with a model prediction confidence safeguard achieves superior results with less data and fewer planning steps compared to uniform sampling experiences.

\section{Related Work}

\subsubsection{Offline Reinforcement Learning (RL)}  In offline RL, agents are trained only from pre-collected datasets, avoiding the risks associated with real-time data collection. One of the main issues in offline RL is the distribution mismatch between the \textit{behavioral (logging) policy} and the \textit{target policy} that is being optimized. 

Fujimoto et al.~\cite{fujimoto19a} tackled this by stabilizing Q-learning to reduce bootstrapping errors caused by this mismatch, specifically through the introduction of a method to limit the overestimation of Q-values. Building on this, Wu et al.~\cite{wu2019behavior} proposed a conservative Q-learning approach known as Behavior-Regularized Actor-Critic (BRAC), which further mitigates Q-value overestimation by incorporating a penalty term that keeps the learned policy close to the behavior policy. Levine et al.~\cite{levine2020offline} emphasize the broader challenges in offline RL, highlighting the necessity of techniques that effectively manage limited and biased datasets. Most prior work in this area focuses only on model-free approaches, exploring algorithms such as the Q-network ensembles~\cite{fujimoto19a,wu2019behavior}, behavioral policy regularization~\cite{jaques2019way,kumar_aviral2019,wu2019behavior}, and implicit Q-learning~\cite{kostrikov2021implicit}.

\subsubsection{Distribution mismatch correction techniques} Various approaches have been developed to mitigate the policy-induced state-action distribution mismatch. Precup et al.~\cite{precup2001off} tackle the problem using products of importance sampling ratios, though this approach suffers from large variance. To correct the distribution mismatch without incurring a large variance, Hallak et al.~\cite{hallak2017consistent} and Liu et al.~\cite{liu2018breaking} propose learning the density ratio between the state distribution of the target policy and sampling distribution directly. 

DualDICE~\cite{nachum2019dualdice} is a relaxation of previous methods and enables learning from multiple unknown behavior policies. They achieved impressive results by using a change of variable technique in the density ratio calculation. GenDICE~\cite{zhang2020gendice} is a generalization of DualDICE, stabilizing estimations in the average reward setting. GradientDICE~\cite{zhang2020gradientdice} outlines that GenDICE is not a convex-concave saddle-point problem in all settings and proposes a provably convergent method under linear function approximation. Despite their differences, all these algorithms use minimax optimizations, allowing them to be combined under the regularized Lagrangians of the same linear problem~\cite{yang2020offpolicy}.

\subsubsection{Model-based offline RL} Model-based RL (MBRL) uses a learned model of the environment to generate additional experiences, thereby enhancing sample efficiency. The Dyna-Q algorithm~\cite{sutton1991dyna} was among the first to integrate model-based techniques with reinforcement learning, combining planning and learning into a unified framework. Over time, MBRL has become a key approach in online reinforcement learning, with several algorithms demonstrating exceptional sample efficiency by incorporating sophisticated dynamics models that closely simulate the environment~\cite{hafner2023mastering,hafner2020mastering}.

However, in offline settings where interactions with the environment are restricted, inaccuracies in the learned models can lead to the generation of unrealistic or 'hallucinated' states within the synthetic experiences. To mitigate this issue, recent methods such as MOPO~\cite{yu2020mopo} and MOReL~\cite{kidambi2020morel} incorporate uncertainty estimates into the model's predictions. This approach helps guide the policy towards safer and more reliable regions of the state space, reducing the risk of overfitting to unreliable synthetic experiences.

In this work, we extend pessimistic offline MBRL approaches by incorporating DICE estimations with a prediction confidence estimation. This combination aims to allow the algorithm to safely explore regions with an estimated divergence between the offline dataset and the candidate policy, improving robustness. Although we explore a simple setting with a Tabular World Model in this work, we believe this research direction has the potential to mitigate overfitting to unreliable synthetic data when applied with more complex world models.

\section{Preliminaries}

In this section, we introduce the theoretical framework and describe the problem setting. We also present the DIstribution Correction Estimation (DICE) algorithm, which is the basis for updating sampling probabilities in SimuDICE.

\subsection{Theoretical framework}

We consider a Markov Decision Process (MDP) \cite{puterman1994markov}, in which the environment is defined by a tuple $\mathcal{M}= \langle S, A, R, T, \mu_0, \gamma \rangle$ where $S$ represents the state space, $A$ is the action space, $R$ is a reward function, $T$ is the transition probability function, $\mu_0$ is the initial state distribution, and $\gamma \in [0, 1)$ is the discount factor.

A policy $\pi$ in an MDP decides what action the agent should take given some state $s$. Formally, it is a mapping $\pi: S \rightarrow \Delta(A)$, where $\pi(s)$ represents the probability distribution over actions A in state $s$. The goal of the agent is to maximize the cumulative expected reward (its return), given by Eq. (\ref{eq:background_policy_evaluation}).

\begin{equation}
\rho(\pi) = \mathbb{E}_{s_0 \sim \mu_0} \left[ \sum_{t=0}^{\infty} \gamma^t R(s_t, a_t) \mid a_t \sim \pi(\cdot \mid s_t) \right]
\label{eq:background_policy_evaluation}
\end{equation}

\noindent
To evaluate the performance of a policy, we define two functions: the Value function $V^{\pi}(s)$, which is the expected return of policy $\pi$ from state $s$ (Eq. \ref{eq:background_value_function}), and the Q-value function $Q^{\pi}(s, a)$, which represents the expected return following a policy $\pi$ starting from state $s$ with action $a$ (Eq. \ref{eq:background_q_function}).

\begin{equation}
V^{\pi}(s) = \mathbb{E}_{\pi} \left[ \sum_{t=0}^{\infty} \gamma^t R(s_t, a_t) \mid s_0 = s \right]
\label{eq:background_value_function}
\end{equation}

\begin{equation}
Q^{\pi}(s, a) = \mathbb{E}_{\pi} \left[ \sum_{t=0}^{\infty} \gamma^t R(s_t, a_t) \mid s_0 = s, a_0 = a \right]
\label{eq:background_q_function}
\end{equation}

\subsubsection{Bellman equation} The Bellman equation provides a recursive definition of the Q-value by decomposing it into immediate reward and the discounted value of the next state-action pair. Eq. (\ref{eq:background_q_value_bellman_equation}) shows how the Bellman equation is applied to the Q-value policy function $Q^{\pi}(s, a)$.

\begin{equation}
Q^{\pi}(s, a) = R(s, a) + \gamma \mathbb{E}_{s' \sim T(s,a), a' \sim \pi(\cdot|s')} \left[ Q^{\pi}(s', a') \right]
\label{eq:background_q_value_bellman_equation}
\end{equation}

\noindent
The Bellman operator $\mathcal{B}^{\pi}$ iteratively applies the Bellman equation to update the Q-values until convergence, leading to a formulation as in Eq. (\ref{eq:background_q_value_bellman_operator}).

\begin{equation}
    \mathcal{B}^{\pi} Q(s, a) = R(s, a) + \gamma \mathbb{E}_{s' \sim T(s,a), a' \sim \pi(\cdot|s')}[Q(s', a')]
    \label{eq:background_q_value_bellman_operator}
\end{equation}

\subsection{Problem setting}

\subsubsection{Offline Reinforcement Learning} The focus of this work is offline RL. Unlike online RL where the agent actively interacts with the environment to gather data and update its policy, offline RL aims to derive the optimal policy $\pi$ from a pre-collected dataset of experiences. Specifically, we assume access to a finite dataset $\mathcal{D} = \left\{ \left(s_0^{(i)}, s^{(i)}, a^{(i)}, r^{(i)}, s'^{(i)} \right) \right\}_{i=1}^{N}$, where $s_0 \sim \mu_0$, $(s^{(i)}, a^{(i)}) \sim d^{\mathcal{D}}$ are samples from an unknown distribution $d^{\mathcal{D}}$, $r^{(i)} \sim R(s^{(i)}, a^{(i)})$, and $s'^{(i)} \sim T(s^{(i)}, a^{(i)})$. 

\subsubsection{Model-Based RL} MBRL involves learning an MDP $\hat{\mathcal{M}}= \langle S, A, \hat{R}, \hat{T}, \hat{\mu_0}, \gamma \rangle$ which uses the learned transitions $\hat{T}$ instead of the true transitions $T$, and the learned reward function $\hat{R}$ instead of the true reward function $R$. In this work, we assume the initial distribution $\mu_0$ is unknown and learned from the data.

\subsection{DualDICE estimation}
\label{subsection:dualdice_estimation}

In this section, we will elaborate on the DICE estimation algorithm used in SimuDICE, namely DualDICE~\cite{nachum2019dualdice}, which is often used for \textit{off-policy evaluation}. They obtained impressive results by reducing the \textit{off-policy evaluation} problem to density ratio estimation and doing a change of variable optimization trick. The policy value can be rewritten using the importance weighing trick (Eq.~\ref{eq:importance_weighting_trick}).

\begin{equation}
    \rho(\pi) = \mathbb{E}_{(s, a) \sim d^\pi} [r(s, a)] = \mathbb{E}_{(s, a) \sim d^\mathcal{D}} \left[ \frac{d^\pi(s, a)}{d^\mathcal{D}(s, a)} r(s, a) \right],
    \label{eq:importance_weighting_trick}
\end{equation}

\noindent
where $d^{\pi}$ is the discounted state visitation distribution (Eq.~\ref{eq:discounted_on-policy_distribution}).

\begin{equation}
    d^\pi(s, a) := (1 - \gamma) \sum_{t=0}^{\infty} \gamma^t \cdot \Pr[s_t = s, a_t = a \mid s_0 \sim \mu_0, \pi]
    \label{eq:discounted_on-policy_distribution}
\end{equation}

\noindent
Eq.(\ref{eq:importance_weighting_trick}) can be rewritten in the offline setting as a weighted average (Eq. \ref{eq:weighted_average_formulation}), reducing the problem to estimating the density ratios (Eq. \ref{eq:just_weights}) for policy correction.

\begin{equation}
    \mathbb{E}_{(s, a) \sim d^\mathcal{D}} \left[ \frac{d^\pi(s, a)}{d^\mathcal{D}(s, a)} r(s, a) \right] = \frac{1}{N} \sum_{i=1}^{N} \frac{d^\pi(s^{(i)}, a^{(i)})}{d^\mathcal{D}(s^{(i)}, a^{(i)})} r^{(i)}
    \label{eq:weighted_average_formulation}
\end{equation}

\begin{equation}
    w_{\pi/\mathcal{D}}(s, a) := \frac{d^\pi(s, a)}{d^\mathcal{D}(s, a)}
    \label{eq:just_weights}
\end{equation}

\noindent
DualDICE~\cite{nachum2019dualdice} optimizes a (bounded) function  \footnote{Note that $\nu$ is a state-action value function, analogous to Q-values, although Q-values were not utilized in this context for completeness and to maintain generality.}: $\nu : S \times A \rightarrow \mathbb{R}$ as in Eq.(~\ref{eq:optimization}).

\begin{equation}
\min_{\nu : S \times A \rightarrow \mathbb{R}} J(\nu) := \frac{1}{2} \mathbb{E}_{(s,a) \sim d^{\mathcal{D}}} \left[ (\nu - \mathcal{B}_0^{\pi} \nu)(s, a)^2 \right] - (1 - \gamma) \mathbb{E}_{s_0 \sim \mu, a_0 \sim \pi(s_0)} \left[ \nu(s_0, a_0) \right]
\label{eq:optimization}
\end{equation}

\noindent
$\mathcal{B}_0^{\pi}$ is used to denote the expected Bellman operator with respect to policy $\pi$ and zero-reward: $\mathcal{B}_0^{\pi} \nu(s, a):= \gamma \mathbb{E}_{s' \sim T(s, a), a' \sim \pi(s')}[\nu(s', a')] $. The authors of DualDICE \cite{nachum2019dualdice} state that the first term alone leads to a trivial solution $\nu \equiv 0$, which is avoided by the second term that ensures $\nu^{*} > 0$. They prove that the Bellman residuals of $\nu^*$ are exactly the desired distribution corrections (Eq. \ref{eq:bellman_residual_equality}).

\begin{equation}
    w_{\pi / \mathcal{D}}(s, a) = (\nu^* - \mathcal{B}_0^{\pi} \nu^*)(s, a).
    \label{eq:bellman_residual_equality}
\end{equation}

\section{SimuDICE}

\begin{figure}[h!]
\centering
\includegraphics[width=0.625\textwidth]{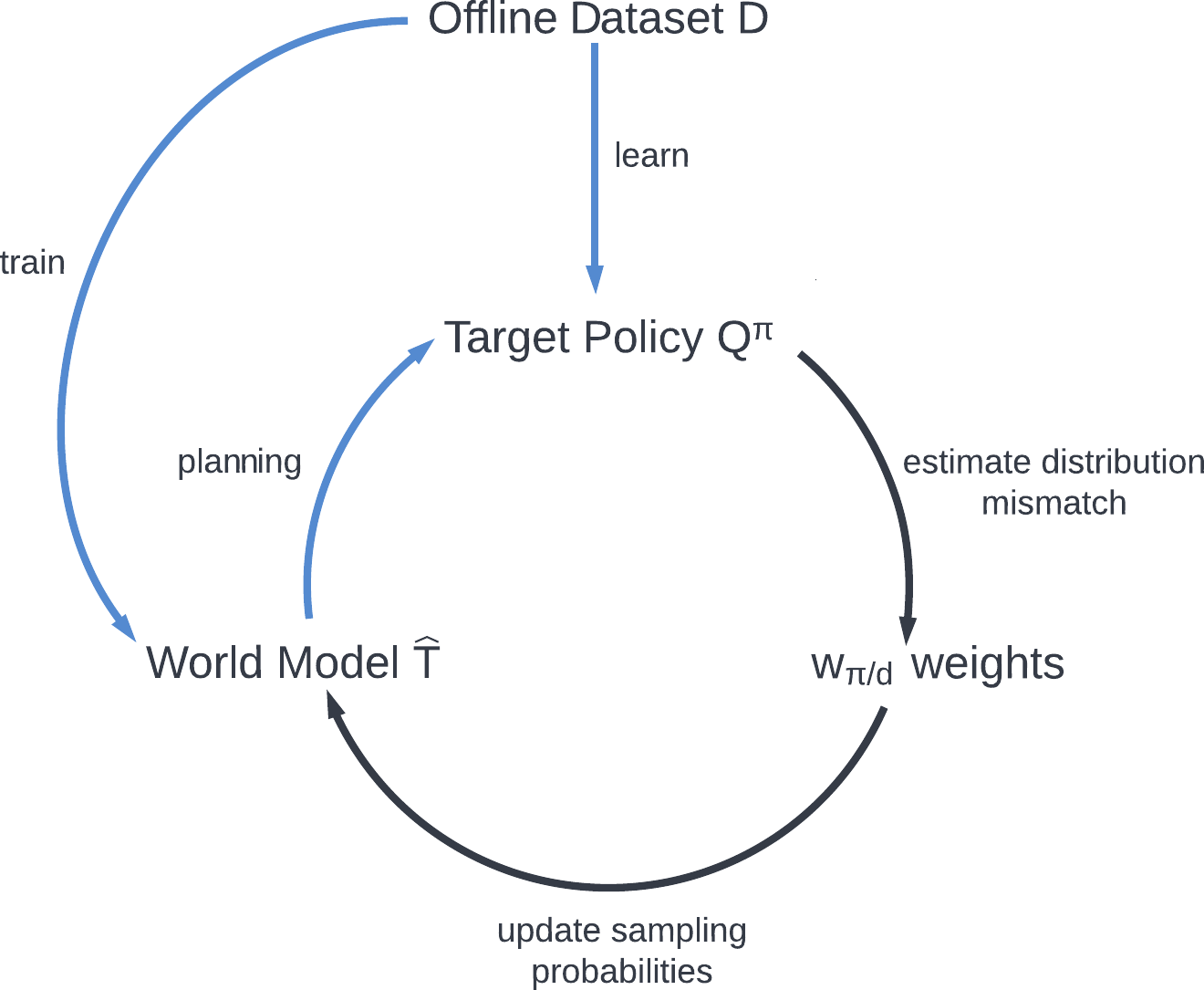}
\caption{The components of SimuDICE and their interactions. Transitions adapted from Dyna-Q \cite{sutton1991dyna} are in blue, while those unique to SimuDICE are depicted in black.} 
\label{fig:algorithm_pipeline}
\end{figure}

For clarity and ease of understanding, we start by presenting an idealized version of SimuDICE, discussing its theoretical foundations. We then proceed to describe the practical implementation used in our experiments. Algorithm~\ref{alg:simudice_high_level} presents the broad framework, while Figure~\ref{fig:algorithm_pipeline} shows an illustration of the algorithm.

\begin{algorithm}
\caption{SimuDICE: Offline Policy Optimization}
\begin{algorithmic}[1]
\Require Dataset $\mathcal{D}$
\State Learn approximate dynamics model $\hat{T} : S \times A \to S$ using $\mathcal{D}$.
\State Initialize estimated model confidence $\mathcal{C}$ using $\mathcal{D}$.
\State Learn an initial target policy $\pi_\text{target}$ using $\mathcal{D}$.
\For{iteration = 1 to \texttt{number\_iterations}}
    \State $w_{\pi/\mathcal{D}} \gets \text{DualDICE}(\mathcal{D}, \pi_\text{target})$\
    \State $\mathcal{P} \gets \text{updateProbabilities}(\mathcal{C}, w_{\pi/\mathcal{D}})$
    \State $\pi_\text{target} \gets \text{planner}(\hat{T}, \mathcal{P}, \pi_\text{target})$
\EndFor
\State \Return $\pi_\text{target}$
\end{algorithmic}
\label{alg:simudice_high_level}
\end{algorithm}

\subsubsection{Learning the world model} 
The first step involves using the offline dataset to learn an approximate dynamics model $\hat{T}(\cdot \mid s, a)$. Usually in literature, this is achieved through maximum likelihood estimator~\cite{deisenroth2011pilco} or other techniques like dynamics modeling~\cite{hafner2023mastering,hafner2020mastering} or diffusion~\cite{alonso2024diffusion,ding2024diffusion}.

In this work, we use for simplicity a Tabular World Model \cite{Sutton1990Integrated}. This memory-based model replicates previously observed experiences by averaging the rewards for each transition and selecting the most frequently observed next state (Algorithm~\ref{alg:world_model_training}). This approach allows the model to handle stochastic environments.

\begin{algorithm}[h!]
\caption{Learning the Tabular World Model}
\begin{algorithmic}[1]
\State \textbf{Initialize} an empty tabular model.
\ForAll{state-action pairs $(s, a)$ in the dataset}
    \State \textbf{Update} the model with the observed next state $s'$ and reward $r$.
\EndFor
\ForAll{state-action pairs $(s, a)$ in the model}
    \State \textbf{Compute} the average reward for $(s, a)$.
    \State \textbf{Determine} the most frequently observed next state $s'$.
\EndFor
\State \textbf{Return} the learned tabular model.
\end{algorithmic}
\label{alg:world_model_training}
\end{algorithm}

\vspace{-0.5cm}
\subsubsection{Learning the initial policy} In this stage, we use the offline pre-collected dataset to derive an initial policy, using Experience Replay \cite{mnih2015humanlevel}, which makes the policy more stable. We apply the Q-learning formula as in Eq. (\ref{eq:q_learing_equation}). 

\begin{equation}
Q(s, a) \leftarrow Q(s, a) + \alpha \left[ r + \gamma \max_{a'} Q(s', a') - Q(s, a) \right],
\label{eq:q_learing_equation}
\end{equation}

\subsubsection{Updating the Sampling Probabilities}

The core of SimuDICE consists of updating the sampling probabilities, a process that involves two steps. Firstly, we calculate the stationary distribution correction, \(w_{\pi/\mathcal{D}}\), using DualDICE~\cite{nachum2019dualdice}.

The second step involves adding a safeguard to guide the model toward `safe experiences.` This safeguard can take various forms, such as incorporating pessimism or other risk-averse strategies. In this work, because our world model cannot predict new experiences, for completeness, we consider the confidence of a prediction, \(\mathcal{C}(s, a)\), as the normalized frequency of occurrences in the dataset.

Let $\mathcal{P}(s, a)$ be the probability that the world model will sample state $s$ and action $a$. We consider $\mathcal{P}(s, a)$ as the normalized sum of the confidence $\mathcal{C}(s, a)$ and the regularized softmax of the $w_{\pi / \mathcal{D}}$ weights. The regularization term \(\lambda\) is introduced to align the scale of confidence predictions with DICE estimations, preventing mode collapse. Additionally, in environments with large state-action spaces, the values of \(w_{\pi / \mathcal{D}}\) become too small, leading to floating-point errors. Below we show the derivation of the likelihood function $\mathcal{P}(s, a)$. First, we define the likelihood \(\mathcal{L}(s, a)\) (Eq.~\ref{eq:L}) of sampling the state-action pair $(s, a)$.

\begin{equation}
    \mathcal{L}(s, a) = \mathcal{C}(s, a) + \frac{e^{w_{\pi / \mathcal{D}}(s, a) \cdot \lambda}}{\sum_{(s', a')} e^{w_{\pi / \mathcal{D}}(s', a') \cdot \lambda}} / \lambda
    \label{eq:L}
\end{equation}

\noindent
The final probability function is obtained by normalizing \(\mathcal{L}(s, a)\) (Eq.~\ref{eq:P}).

\begin{equation}
    \mathcal{P}(s, a) = \frac{\mathcal{L}(s, a)}{\sum_{(s', a')} \mathcal{L}(s', a')}
    \label{eq:P}
\end{equation}

\subsubsection{Planning} The planning phase of the algorithm improves the policy using synthetically generated data, enabling the agent to learn from a more diverse range of experiences. The policy is updated using the same method as experiences from the offline dataset. In SimuDICE, experiences $(s, a)$ are sampled using the World model $\hat{T}$ with different probabilities, calculated based on how likely is the experience to be encountered by running the policy and how confident is the world model on the prediction. The world model predicts the subsequent reward and next state $(r, s')$ and updates the Q-values using Eq. \ref{eq:q_learing_equation}.  A pseudocode of how planning works can be seen in Algorithm~\ref{alg:planner}.

\begin{algorithm}
\caption{Planner}
\begin{algorithmic}[1]
\Require Dynamics model $\hat{T}$, Probability distribution $\mathcal{P}$, Current policy $\pi$
\For{iteration = 1 to \texttt{planning\_iterations}}
    \State Sample state-action pair $(s, a)$ using $\mathcal{P}$
    \State $(s', r) \gets \hat{\mathcal{T}}(s, a)$
    \State Update policy $\pi$ using the observed transition $(s, a, s', r)$
\EndFor
\State \Return Updated policy $\pi$
\end{algorithmic}
\label{alg:planner}
\end{algorithm}

\section{Experiments}
\label{sec:experiments}

Our experiments aim to answer the following questions: 1) how do the quality (value) and size (number of experiences collected) of the behavior policy impact the policy learned by SimuDICE; 2) how does the policy derived by SimuDICE compare with other algorithms across different off-policy datasets; and 3) what effect do various components and parameters have on SimuDICE's performance?

\begin{figure}
\includegraphics[width=\textwidth]{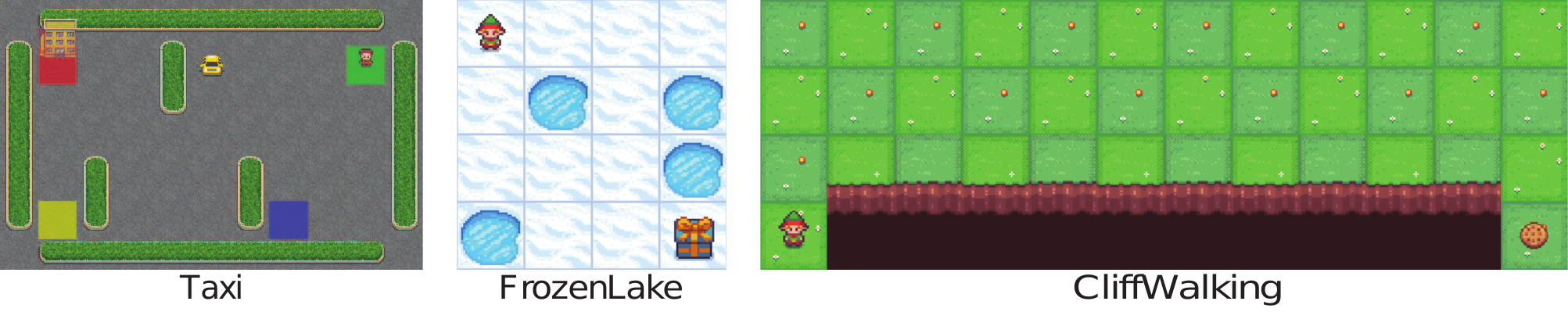}
\caption{Illustration of the suite of tasks considered in this work. These tasks require the RL agent to learn to navigate grid environments to accomplish certain tasks.} 
\label{fig:environment_images}
\end{figure}

To answer the above questions, we consider commonly studied benchmark tasks from the Gymnasium Library~\cite{towers_gymnasium_2023}. Our experimental setup follows previous work~\cite{fujimoto19a,kumar2019stabilizing,wu2019behavior}. We focus on \textit{Toy Text} environments: \textit{Taxi}, \textit{FrozenLake}, and \textit{CliffWalking}, as shown in Figure~\ref{fig:environment_images}. For each environment, we consider three distinct logged datasets. These datasets are collected following Wu et al.~\cite{wu2019behavior}, with 500 timesteps of environment interaction each. First, we partially train a policy ($\pi_{p}$) to achieve values of approximately $0.1$, $0$, and $-2.38$, respectively. We generate three noisy $\pi_{p}$ variants using epsilon-greedy strategies ($\epsilon$ = 0.1, 0.4, 0.7), introducing different noise magnitudes. The evaluation was performed on the \textit{Taxi} environment using the 9 datasets collected from the other environments.

We evaluate the learned policies by performing rollouts in the (real) environment, strictly for evaluation purposes. These rollouts are not available to the algorithm and are not used for any learning. This evaluation protocol is consistent with prior work~\cite{fujimoto19a,kumar2019stabilizing,wu2019behavior}. We report results as average per-step reward over 500 plays and 20 seeds, using identical hyperparameters across all environments. Unless stated otherwise, we used the following hyperparameters: $\alpha$: \texttt{0.1}, $\gamma$: \texttt{0.99}, \texttt{10} planning steps, \texttt{1} iteration, regularization parameter ($\lambda$): \texttt{1000}, play episodes: \texttt{500}, and environment steps: \texttt{100}.

\vspace{-0.5cm}
\subsection{Algorithm comparisons}
\label{subsec:algorithm comparisons}

We compare SimuDICE with two other methods: offline Q-learning and a variant of SimuDICE that uses uniform sampling probabilities, which we refer to as offline Dyna-Q. To evaluate their effectiveness across different planning scenarios, we assess both SimuDICE and offline Dyna-Q using 10 and 20 planning steps.

\begin{figure}[!ht]
\centering
\makebox[\textwidth][c]{%
    \includegraphics[width=1\textwidth]{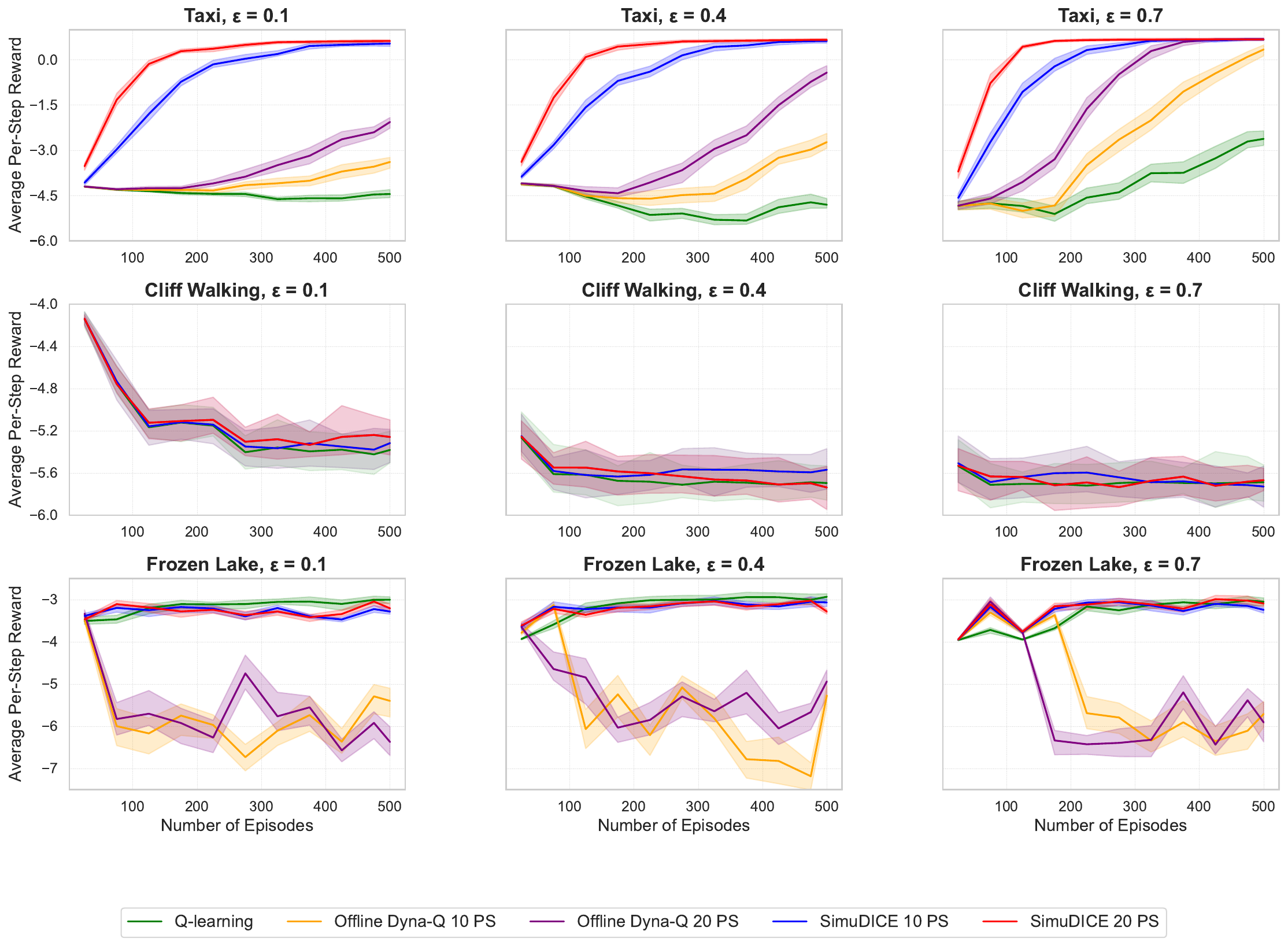}
}
\caption{Comparison of algorithm performance in discrete tabular environments: \textit{Taxi}, \textit{CliffWalking}, and \textit{FrozenLake}, under varying epsilon-greedy data collection policies ($\epsilon = 0.1, 0.4, 0.7$). Each plot shows the average per-step reward as a function of the number of trajectories in the offline data. The results are averaged over 500 episodes and 5 different random seeds. The shaded regions represent the variance across the different seeds. PS represents the number of planning steps.}
\label{fig:environment_result}
\end{figure}

Figure~\ref{fig:environment_result} shows that SimuDICE consistently outperforms or matches the performance of the other algorithms across various settings. In particular, in the Taxi environment, which is considered more challenging due to its larger state-action space and the diversity of actions and penalties, SimuDICE significantly outperforms the other algorithms. Even with only 10 planning steps, SimuDICE's performance often surpasses the performance of the 20-step offline Dyna-Q variant, with this difference being more pronounced in environments where data was collected with smaller exploration terms (lower epsilon-greedy values).

\noindent
In the CliffWalking environment, the performance of algorithms shows no significant difference, with all being within the same variance range.

In the FrozenLake environment, both offline Q-learning and SimuDICE perform similarly, achieving comparable average per-step rewards with no noticeable differences. The key observation is that the offline Dyna-Q method consistently underperforms compared to the others, regardless of the offline data quality.

\subsection{Ablation study}
\label{subsec:ablation_study}

\noindent
In this section, we conduct an ablation study to evaluate the effect of various parameters on the performance of SimuDICE. Our analysis focuses only on the \textit{Taxi} environment, as it represents the most complex scenario out of the three. For this ablation study, we have changed $\alpha$ to $0.05$ to visualize the results better.

\subsubsection{Planning Steps:} \textit{How does the number of planning steps affect the model's performance?} 

\begin{figure}[h!]
\centering
\includegraphics[width=\textwidth]{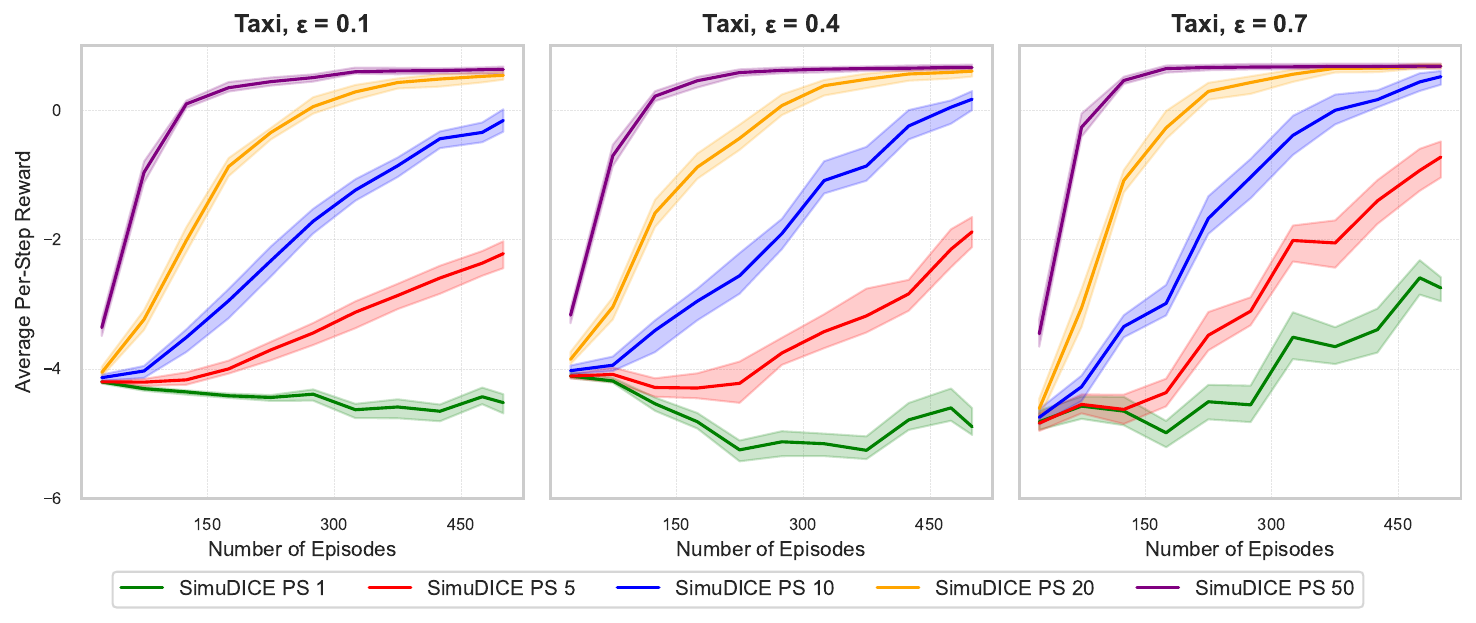}
\caption{Impact of the number of planning steps on the average per-step reward under different epsilon-greedy data collection policies, with varying epsilon values.}
\label{fig:ablation_planning_steps_results}
\end{figure}

We carry out an experimental evaluation to determine how different numbers of planning steps affect the agent's performance. Figure \ref{fig:ablation_planning_steps_results} shows that while the number of planning steps improves the performance, the relationship is not linear. The improvement is particularly evident in low-data cases, with the difference starting to decrease with the number of experiences in the offline dataset.

\vspace{-0.75cm}
\subsubsection{Different Sampling Probabilities Formulas:}  \textit{How does the algorithm's performance change when we alter the method for estimating sampling probabilities?}  

We compare three different formulas for altering the sampling probabilities using the \( w_{\pi / \mathcal{D}} \) weights and model confidence estimation. This comparison assesses how effectively these approaches prioritize `valuable` synthetic experiences for the world model. To ensure normalization, likelihoods are converted into probabilities by dividing each by their total sum.

\begin{itemize}
    \item \textbf{Formula 1}: The default in SimuDICE (Eq.~\ref{eq:function1}).
    \item \textbf{Formula 2}: Similar to Formula 1 but discourages experiences with distribution shifts instead of encouraging them (Eq.~\ref{eq:function2}).
    \item \textbf{Formula 3}: Applies lambda-regularized DICE estimations \( w_{\pi / \mathcal{D}} \) over a uniform sampling model, ignoring model confidence (Eq.~\ref{eq:function3}).
\end{itemize}

\begin{figure}[h!]
\centering
\includegraphics[width=\textwidth]{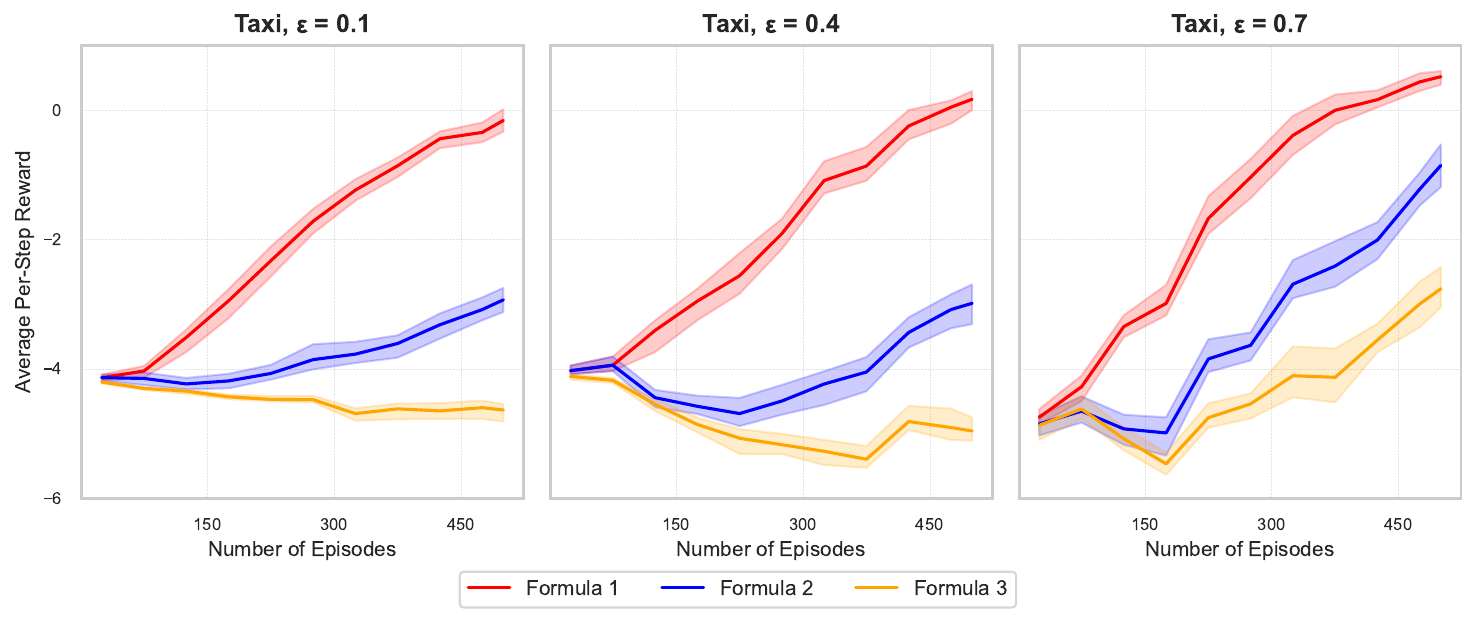}
\caption{Comparison of average per-step rewards achieved by SimuDICE using different sampling formulas, across different epsilon-greedy offline dataset collection policies.}
\label{fig:ablation_formulas_results}
\end{figure}

\noindent
Figure \ref{fig:ablation_formulas_results} shows that the formula used in SimuDICE outperforms others under varying data qualities. However, when the \textit{target policy} is close to the \textit{behavioral policy} used for data collection, alternative sampling methods may outperform it. Specifically, the SimuDICE formula excels in scenarios with diverse data but yields inferior results when the data lacks diversity and is already close to the desired distribution).

\begin{align}
  \mathcal{L}_1(s, a) &= \mathcal{C}(s, a) + \frac{e^{w_{\pi / \mathcal{D}}(s, a) \cdot \lambda}}{\lambda \sum_{(s', a')} e^{w_{\pi / \mathcal{D}}(s', a') \cdot \lambda}} 
  \label{eq:function1} \\ 
  \mathcal{L}_2(s, a) &= \mathcal{C}(s, a) - \frac{e^{w_{\pi / \mathcal{D}}(s, a) \cdot \lambda}}{\lambda \sum_{(s', a')} e^{w_{\pi / \mathcal{D}}(s', a') \cdot \lambda}} 
  \label{eq:function2} 
  \\
  \mathcal{L}_3(s, a) &= \frac{1}{S \cdot A} + \frac{e^{w_{\pi / \mathcal{D}}(s, a) \cdot \lambda}}{\lambda \sum_{(s', a')} e^{w_{\pi / \mathcal{D}}(s', a') \cdot \lambda}} 
  \label{eq:function3}
\end{align}

\vspace{-0.5cm}

\vspace{-0.5cm}
\subsubsection{Number of iterations:} \textit{How does the number of iterations (the number of updates to the sampling probabilities) change the performance of the algorithm?} 

\begin{figure}[h!]
\centering
\includegraphics[width=\textwidth]{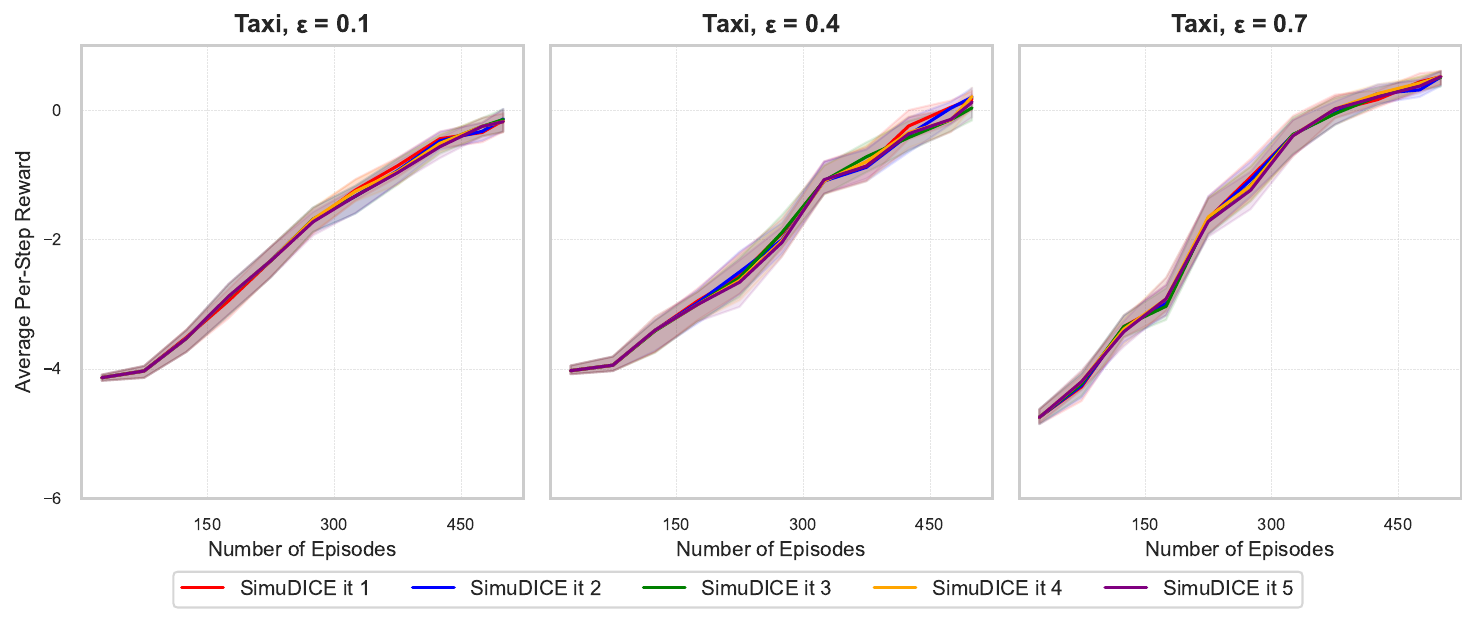}
\caption{Effect of iteration number on the average per-step reward achieved across different epsilon-greedy offline dataset collection policies by SimuDICE.}
\label{fig:ablation_iterations_results}
\end{figure}

To verify the effectiveness of the number of iterations (i.e., the frequency of updating the sampling probabilities) on the agent's performance, we conducted a comparative analysis using the SimuDICE with 10 planning steps.

Figure \ref{fig:ablation_iterations_results} shows that varying the number of iterations has a negligible effect on the performance of SimuDICE in the Taxi environment.

\section{Discussion}

This work introduced SimuDICE, a framework for policy optimization in offline reinforcement learning, and assessed its effectiveness across various scenarios.

\noindent
SimuDICE addresses the \textit{data needs} and the \textit{state-action distribution mismatch}. In this section, we discuss the findings, their implications, and their limitations.

\subsubsection{Key Findings and Implications} Based on the experimental results, we derive the following key findings and implications of the proposed method:

\begin{itemize} 
    \item \textbf{Improved Sample Efficiency:} Our experiments show that SimuDICE achieves greater sample efficiency compared to uniform sampling experiences during the planning stage and vanilla Q-learning. This advantage is particularly evident in data-rich environments where the collected data diverges significantly from the optimal policy. We attribute this improvement to the adjustment of sampling probabilities using DICE estimations. 
    \item \textbf{Distribution Mismatch Correction:}  Our results show that in addition to SimuDICE being more sample-efficient than the methods we compared it with, it also requires fewer planning steps to achieve comparable or superior policies. This suggests that we can achieve equal or better policies while generating fewer synthetic experiences using the world model, which implicitly reduces the risk of `hallucination`. While the DICE estimations play an important role in this improvement, it is worth mentioning that the confidence prediction estimation was added for completeness as a safeguard to balance exploitation-exploration for these experiences.
    \item \textbf{Objective Mismatch:} Our ablation study reveals no significant performance gain from updating the sampling probabilities multiple times during a run. We believe this may be due to the simplicity of the environments used in our experiments, as the approach was originally theorized for more complex settings~\cite{eysenbach2021mismatched,lambert2020objective}. Additionally, the basic nature of the components, such as the world model, may have influenced these results.
\end{itemize}

\subsubsection{Limitations} In this part, we outline the limitations of this study and indicate some possible future work directions.

\begin{itemize}
    \item \textbf{Simple environments:} While SimuDICE showed promising results in deterministic grid-world environments, their simplicity limits the generalizability of the findings. Therefore, future research should focus on evaluating SimuDICE in more complex settings to assess its effectiveness better.
    \item \textbf{Sensitivity to sample probabilities formula:} In the ablation study, we show that the performance of SimuDICE is affected by different variations of the sampling probability formula. Given the limited number of scenarios the algorithm was tested on, a more extensive evaluation across a wider range of environments is necessary to assess its effectiveness.
    \item \textbf{Comparison with other algorithms:} This study conducted a comparison between SimuDICE, vanilla offline Q-learning, and offline Dyna-Q~\cite{sutton1991dyna}, where the latter is effectively SimuDICE with uniform sampling probabilities. While this comparison gives an intuition of its improvements, a comparison with other algorithms might provide more insights into its performance. 
\end{itemize}

\section{Conclusion}
We have presented SimuDICE, a framework for optimizing policies in model-based offline reinforcement learning by adjusting the world model's sampling probabilities using DualDICE estimation and the estimated prediction confidence. The main innovation of SimuDICE lies in its ability to correct the state-action distribution mismatch between the \textit{behavior policy} and \textit{target policy} through a bi-objective optimization that balances experience realism and diversity, thereby preventing mode collapse and the generation of hallucinated states.

Our experiments show that modifying sampling probabilities offers advantages over uniform sampling, achieving similar average per-step rewards with fewer pre-collected experiences and planning steps. SimuDICE also demonstrates greater robustness to variations in data collection policies.

Future work includes (1) incorporating a world model that can generate novel experiences, using a more stable and robust DICE estimator, and implementing a policy that can handle continuous state-action spaces; (2) further exploring the impact of altered sampling probabilities on the stability and robustness of SimuDICE across various data collection policies; and (3) evaluating the algorithm's performance in more complex, continuous environments where greater divergence between the behavioral and target policies is present.

\begin{credits}
\subsubsection{\ackname} S. Bongers and F. Oliehoek are supported by the Mercury Machine Learning Lab, a collaboration between TU Delft, UvA, and booking.com.

\end{credits}

\clearpage

\bibliographystyle{splncs04}

\end{document}